\documentclass[lettersize,journal]{IEEEtran}
\usepackage{amsmath,amsfonts}
\usepackage{algorithmic}
\usepackage{array}
\usepackage[caption=false,font=normalsize,labelfont=sf,textfont=sf]{subfig}
\usepackage{textcomp}
\usepackage{pifont}
\usepackage{stfloats}
\usepackage{url}
\usepackage{verbatim}
\usepackage{graphicx}
\usepackage{xcolor}
\usepackage{colortbl}
\usepackage{multirow}
\usepackage{orcidlink}
\usepackage{xcolor} 
\usepackage{amssymb} 
\usepackage{pifont} 
\usepackage{bm}   
\usepackage{booktabs}
    
\def\BibTeX{{\rm B\kern-.05em{\sc i\kern-.025em b}\kern-.08em
    T\kern-.1667em\lower.7ex\hbox{E}\kern-.125emX}}
\usepackage{balance}

\begin{document}
\title{Exploring Partial Multi-Label Learning via Integrating Semantic Co-occurrence Knowledge}

\author{Xin Wu\orcidlink{0009-0008-2292-4534},  Fei Teng\IEEEauthorrefmark{1}\orcidlink{0000-0001-9535-7245}, Yue Feng\orcidlink{0009-0007-3946-0185}, Kaibo Shi\orcidlink{0000-0002-9863-9229}, Zhuosheng Lin\orcidlink{0000-0001-5963-8525}, Ji Zhang\orcidlink{0000-0001-6949-3673} and James Wang\orcidlink{0000-0002-8497-9943}
\thanks{
This work was supported by the National Natural Science Foundation of China (No.62272398), Sichuan Science and Technology Program (No.2024NSFJQ0019). \emph{(Corresponding author: Fei Teng)}

Xin Wu, Fei Teng, and Ji Zhang are with the School of Computing and Artificial Intelligence, Southwest Jiaotong University, Chengdu, Sichuan 611756, China (e-mail: wu1351658806@163.com; fteng@swjtu.edu.cn; jizhang.jim@gmail.com). Fei Teng is also with the Engineering Research Center of Sustainable Urban Intelligent Transportation, Ministry of Education, Chengdu, Sichuan 611756, China.

Yue Feng and James Wang are with the School of Engineering, Swinburne University of Technology, Hawthorn, Victoria 3122, Australia (e-mail: J002443@wyu.edu.cn; jawang@swin.edu.au). Yue Feng and Zhuosheng Lin are with the School of Electronic and Information Engineering, Wuyi University, Jiangmen, Guangdong 510006, China (e-mail: zhuoshenglin@wyu.edu.cn).

Kaibo Shi is the College of Electrical Engineering, Sichuan University, Chengdu, Sichuan 610065, China, and also with the School of Computer Science, Chengdu University, Chengdu, Sichuan 610106, China (e-mail: skbs111@163.com)
}}

\markboth{Journal of \LaTeX\ Class Files,~Vol.~18, No.~9, September~2024}%
{How to Use the IEEEtran \LaTeX \ Templates}

\maketitle

\begin{abstract}
Partial multi-label learning aims to extract knowledge from incompletely annotated data, which includes known correct labels, known incorrect labels, and unknown labels. The core challenge lies in accurately identifying the ambiguous relationships between labels and instances. In this paper, we emphasize that matching co-occurrence patterns between labels and instances is key to addressing this challenge. To this end, we propose the Semantic Co-occurrence Insight Network (SCINet), a novel and practical framework for partial multi-label learning. Specifically, SCINet introduces a bi-dominant prompter module, which leverages an off-the-shelf multimodal model to capture text-image correlations and enhance semantic alignment. To reinforce instance-label interdependencies, we develop a cross-modality fusion module that jointly models inter-label correlations, inter-instance relationships, and co-occurrence patterns across instance-label assignments. Moreover, we propose an intrinsic semantic augmentation strategy that enhances the model’s understanding of intrinsic data semantics by applying diverse image transformations, thereby fostering a synergistic relationship between label confidence and sample difficulty. Extensive experiments on four widely-used benchmark datasets demonstrate that SCINet surpasses state-of-the-art methods.

\end{abstract}


\section{Introduction}

\begin{figure}
\centering
\includegraphics[width=0.48\textwidth]{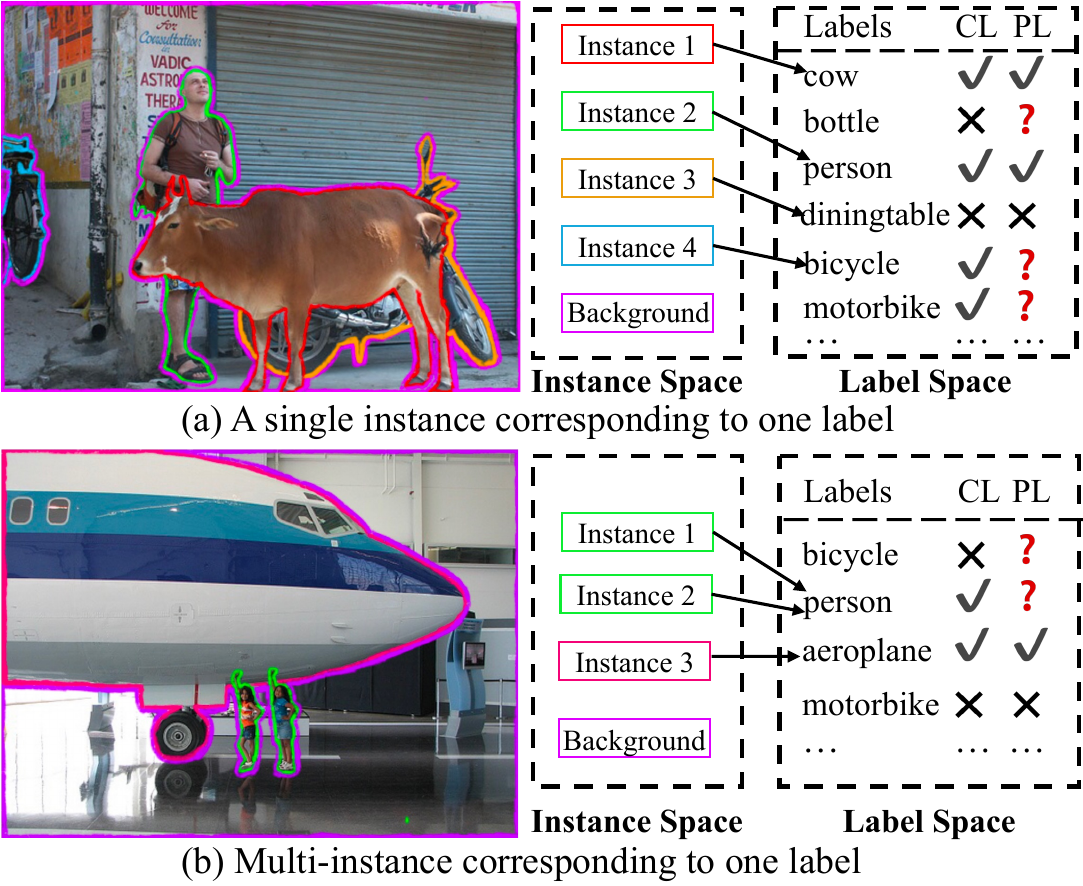}
\caption{Comparison of the matching patterns between instance and label space. "(a)" shows single instance corresponding to one label, while "(b)" depicts multi-instance corresponding to one label. Both "{\normalsize \checkmark}" and "{\normalsize \textbf{\texttimes}}" labels are considered known labels, whereas the "{\normalsize \textbf{\textcolor{red}{?}}}" label is regarded as an unknown label. CL denotes complete label, while PL signifies partial label.} \label{figure1}
\end{figure}

\IEEEPARstart{M}{ulti-label} learning has demonstrated tremendous potential in various fields. However, due to the high cost of labeling and the subjectivity of annotators, real-world datasets often suffer from incomplete and noisy labels. This challenge has spurred the exploration of partial multi-label learning (PML) methods, which aim to address these issues more effectively. Consequently, driven by this research need, PML has garnered vibrant attention in machine learning \cite{67,68}. It represents a new paradigm for multi-label recognition (MLR) and has been widely applied in micro-video classification \cite{44}, image recognition \cite{45}. In these applications, the training dataset typically contains only partial labels rather than complete ones, which better reflects the complexity of real-world data. Unlike standard partial label learning, which selects the true label from a candidate set, our setting focuses on completing the missing entries in the label matrix based on known positive and negative annotations. For instance, in a scenario where one instance corresponds to a single label, the model only needs to identify the correct label for that specific instance (Fig. \ref{figure1} (a)). Conversely, in a setting where multiple instances correspond to a single label, the model must extract features from several instances to determine the common label (Fig. \ref{figure1} (b)).

The key to addressing PML tasks lies in deeply mining discriminative semantic features and constructing a collaborative representation model to fully leverage both label and instance information. A straightforward approach involves leveraging matrix factorization with auxiliary information to accelerate label completion and enhance prediction performance \cite{62}. However, such works suffer from performance limitations due to their neglect of high-order correlations among labels. This inherent constraint has motivated extensive research on capturing label correlations and exploiting data structural information through various mechanisms. For instance, Xie et al. \cite{43} proposed a partial label framework that assigns confidence scores to candidate labels while utilizing data structural information to jointly optimize the classification model and refine confidence estimations for ground-truth labels. Concurrently, semantic-aware modules \cite{31,32} and loss calibration techniques \cite{14} have been widely adopted in partial multi-label frameworks due to their competitive performance in exploring label correlations.

Nevertheless, most existing methods overlook the intrinsic associations between semantic labels and local image instances, rendering models insufficient in capturing fine-grained correspondences between localized features and specific labels. This critical deficiency may compromise model generalization in complex scenarios, particularly under challenging conditions involving occlusions, background clutter, or high inter-class similarity. To address this, emerging research endeavors increasingly focus on incorporating fine-grained association modeling between local features and labels within partial multi-label frameworks. Such approaches aim to enhance discriminative feature extraction and classification robustness even when confronted with noisy labels and complex backgrounds. Representative works include Zhang et al. \cite{51}, who developed a reliable label-guided two-stage approach demonstrating effectiveness in incomplete supervision scenarios, and Lyu et al. \cite{48}, whose probabilistic graph matching mechanism significantly improved the learning process of accurate classifiers from noisy label data.

To date, visual-language pre-trained models have exhibited outstanding performance across various visual tasks. These models, such as the CLIP \cite{61},  pre-trained on a vast corpus of image-text pairs, benefit from large-scale pre-training and effectively bridge the gap between vision and language, providing rich prior knowledge for downstream tasks. In the context of PML, these pre-trained models have demonstrated significant advantages. Even when annotations are available for only a subset of categories, they can leverage their generalization capabilities and existing knowledge of visual-language associations to accurately infer unseen categories, thereby enhancing recognition accuracy and robustness. Sun et al. \cite{7} utilized the strong alignment capability of pre-trained visual-language models by introducing a pair of learnable positive and negative prompts, along with a category-specific region feature aggregation method, addressing how to train models and improve recognition accuracy under limited annotations effectively. Ding et al. \cite{8} particularly highlighted the importance of leveraging structured semantic relationships between labels to address the lack of supervision in situations of incomplete labeling. By introducing cross-modal prompters and a semantic association module, their method more effectively extracts and utilizes the rich knowledge embedded in pre-trained visual-language models, such as CLIP, resulting in performance improvements on MLR tasks beyond existing approaches. In addition, Wang et al. \cite{52} proposed a hierarchical semantic prompting network using CLIP for hierarchical prompt learning, aimed at multi-label classification tasks with a single positive label. As a result, we have adopted techniques based on multi-modal large language models to tackle PML tasks, thereby enhancing performance and accuracy.

Although the aforementioned methods have achieved satisfactory results to a certain extent, their effectiveness remains constrained in addressing the challenge of insufficient supervision, particularly regarding inadequate exploitation of supervisory signals among instances. In reality, inter-instance relationships can provide crucial clues for understanding and predicting attributes of unlabeled instances. Taking image data as an example, instances within images typically do not exist in isolation but exhibit intrinsic interconnections. Such relational patterns offer additional contextual information that facilitates more accurate target recognition under incomplete label supervision. Moreover, large-scale cross-modal datasets inherently contain rich relational information among instances. While most current multi-modal representation learning approaches have yet to fully exploit these patterns and interactions in PML tasks, they demonstrate significant potential for enhancing model capabilities in comprehending and modeling instance-level associations within complex scenes.

Based on the aforementioned issues, we propose a novel approach to address these challenges. This method begins by deeply excavating the discriminative semantic features embedded within the image context. Subsequently, it constructs a collaborative representation model of labels and instances in a high-dimensional data space, thereby achieving precise mapping across a cross-modal semantic space. Specifically, to tackle the challenges posed by incomplete labels, we focus on matching multiple semantic co-occurrence relationships between labels and instances. By thoroughly analyzing their interaction patterns, the model is capable not only of identifying existing semantic labels within the data but also of inferring unseen labels based on the associations among known instances. Consequently, this significantly enhances the overall performance of PML tasks in complex scenarios and large-scale cross-modal environments, offering a new perspective on addressing the challenge of partial labeling.

The critical contributions of this study include:
\begin{itemize}
    \item We propose a novel network, which comprehensively considers the co-occurrence possibilities among labels, among instances, and across different instance–label assignments, thereby effectively guiding the alignment between instances and labels.
     \item The cross-modality fusion module is designed to optimize label confidence by deeply integrating textual and visual data. This module not only focuses on local similarities between samples but also takes into account the global correlations between labels. 
     \item The intrinsic semantic augmentation strategy improves comprehension of data characteristics and semantics by employing threefold image transformations, fostering a synergistic connection between label confidence and sample complexity. This strategy guarantees performance optimization despite the presence of partial labels.
     \item Extensive experiments and analyses on four benchmark datasets demonstrate that the proposed SCINet outperforms state-of-the-art methods.
\end{itemize}


\section{Related Work}
\label{relatedwork}
This section provides a concise overview of previous research on multi-label image classification tasks, focusing on MLR with comprehensive annotations, PML, and multi-modal representation learning in visual tasks.

\subsection{Multi-label Learning}
Multi-label learning remains a pivotal yet highly debated topic in machine learning research \cite{60}. A prevalent approach involves decomposing multi-label learning problems into numerous independent binary classification tasks \cite{16}. However, this methodology not only neglects inter-label correlations but also incurs linearly scaled prediction costs as the label space expands. To address challenges posed by high-dimensional output spaces, alternative studies have attempted to transform the problem into hierarchical multi-class classification subtasks. Specifically, Wei et al. \cite{15} achieved multi-label image classification by processing arbitrary numbers of object segmentation hypotheses through a shared CNN for feature extraction, followed by max-pooling aggregation of hypothesis-specific outputs. Meanwhile, Pham et al. \cite{77} conceptualized images as "bags" containing multiple instances and proposed a dynamic programming-based multi-instance multi-label framework that enables efficient inference from bag-level to instance-level labels. Despite the advancements achieved by the aforementioned methods, they universally overlook latent interdependencies and correlations among labels, consequently exhibiting inherent limitations in capturing global structural information critical for multi-label tasks.

In recent years, an increasing number of studies have focused on exploring the relationships between labels. These studies have significantly advanced the field by leveraging the dependencies among labels to improve classification accuracy. Wang et al. \cite{80} modeled the semantic correlations between images and labels as well as the co-occurrence dependencies among labels by learning a joint image-label embedding space. Furthermore, graph convolutional networks (GCNs) have been increasingly applied to multi-label learning problems \cite{81}, utilizing label correlations to enhance learning performance. Chen et al. \cite{1} constructed a directed graph among labels and employed GCNs to map label embeddings (e.g., word embeddings) into interdependent object classifiers, thereby capturing label dependencies to improve recognition performance. In summary, while previous methods have contributed to the advancement of multi-label learning to varying degrees, they have largely overlooked the bridging role of implicit semantics between observed labels and visual content. In our proposed approach, we systematically mine fine-grained semantic information from known labels and effectively capture the underlying deep semantic associations among labels, instances, and between labels and instances by reinforcing cross-domain interaction mechanisms.

\subsection{Partial Multi-Label Learning}

In PML, the objective is to learn from a label matrix with missing entries, where available annotations include both known positive and known negative labels. This scenario is more challenging than traditional multi-label learning with complete label assignments. Specifically, our goal is to recover the statuses of unobserved labels by leveraging semantic co-occurrence, distinct from the disambiguation task in standard partial label learning. Formally, given a training dataset $\mathcal{D} = \{(\mathbf{X}_i, Y_i)\}_{i=1}^{N}$, where $\mathbf{X}_i \in \mathbb{R}^d$ is a d-dimensional feature vector representing the $i$-th image. The label space $\mathcal{Y}$ for the $i$-th instance is partitioned into three disjoint subsets: the set of verified positive labels $\mathcal{Y}^+_i$, the set of verified negative labels $\mathcal{Y}^-_i$, and the set of unobserved (missing) labels $\mathcal{Y}^U_i$. The goal is to leverage the partial supervision from $\mathcal{Y}^+_i$ and $\mathcal{Y}^-_i$ to accurately infer the true status of labels in $\mathcal{Y}^U_i$. Early studies \cite{84} often treated missing labels as negative labels by default. However, this simplistic approach can introduce significant label bias, as it may erroneously classify a substantial number of true positive labels as negative, thereby adversely affecting the model's learning performance. To address this issue, several methods have been proposed. For instance, some approaches leverage instance similarity, label co-occurrence information, and low-rank regularization to recover missing labels through matrix completion techniques \cite{85}. Kapoor et al.  \cite{86} treated missing labels as latent variables within a probabilistic framework and updated them via Bayesian inference. Additionally, Wu et al. \cite{87} introduced a third state for missing labels and reconstructed the complete label information by enforcing consistency constraints and label assignment smoothness constraints. It is noteworthy that some earlier works \cite{88} still adhered to the strategy of treating unknown labels as negative, while the approach of decomposing the multi-label task into multiple independent binary classification problems overlooked the inherent dependencies among labels.

In light of the aforementioned limitations, subsequent research has increasingly focused on leveraging label correlations to supplement unknown label information, employing methods such as low-rank regularization, label dependency propagation, and hybrid graph modeling \cite{62}. However, these approaches typically require loading the entire training dataset into memory, making them difficult to integrate with fine-tuning strategies for deep neural networks based on mini-batch training. To address this, recent studies have shifted their attention to designing novel loss functions and optimization strategies. For instance, Durand et al. \cite{45} proposed a generalization of the standard binary cross-entropy loss by incorporating label proportion information. Huynh and Elhamifar \cite{98} introduced regularization to the loss function by considering statistical co-occurrence and image-level feature similarity. Pu et al. \cite{32} enhanced samples by generating composite labels through the mixing of category-specific features in the feature space. Meanwhile, Xie et al. \cite{19} utilized structural information from both feature and label spaces to assign confidence scores to potential labels, while Wang et al. \cite{20} guided label disambiguation using interior-point methods and gradient boosting algorithms. He et al. \cite{21} designed a soft labeling threshold operator to narrow the gap between candidate labels and ground truth labels. On the other hand, Guan et al. \cite{22} proposed a semi-supervised PML method that combines partially labeled data with unlabeled data, capturing intrinsic label relationships through low-rank and manifold constraints. To address the issue of significant dependencies among labels, Chen et al. \cite{31} inferred missing labels by leveraging known label information within and across images, leading to the development of the HST method \cite{60}, which is based on category-specific feature-prototype similarity and differential threshold learning. Although HST incorporates adaptive threshold learning, it still relies on modeling inter-class similarities, which may vary across datasets, potentially affecting generalization performance. These methods primarily rely on co-occurrence information within datasets, leading to suboptimal inference when labels are sparse or unstable. In contrast to existing approaches, our method leverages pre-trained multi-modal models, enabling effective reasoning in low-label scenarios by incorporating prior knowledge. To tackle the issue of incomplete label annotations, we introduce an intrinsic semantic augmentation strategy to ensure robust performance.

\subsection{Multi-modal Semantic Fusion}
Multi-modal representation learning plays a pivotal role in downstream visual processing tasks \cite{100}. Taking the CLIP model as an example, it employs an image-text contrastive learning objective, optimizing the strategy to minimize the distance between matching image-text pairs in the embedding space while maximizing the distance between non-matching pairs, thereby effectively capturing the correspondence between images and texts \cite{23}. Although CLIP was initially designed for single-label multi-class classification tasks, its remarkable generalization performance across various visual tasks has sparked extensive exploration into the rich interactive information between vision and language within the research community \cite{24}.

Based on this context, researchers have proposed various innovative methods to explore further and leverage the semantic relationships between vision and language. For instance, Wang et al. \cite{52} developed a novel hierarchical semantic prompting network that dynamically captures the hierarchical semantic relationships between images and text, effectively enhancing the model's ability to understand complex semantic structures. Concurrently, He et al. \cite{59} introduced an open-vocabulary multi-label classification framework based on multi-modal knowledge transfer, aiming to address the issue of unseen labels in real-world MLR systems. Additionally, Wang et al. \cite{99} designed a multi-modal context prompting learning method, which incorporates learnable context prompts in both image and text modalities and utilizes coupling functions to facilitate interactive connections between modal prompts, thereby strengthening the correspondence between images and labels and improving the accuracy of MLR. Liu et al. \cite{101} proposed a novel multi-label few-shot image classification framework based on paired feature enhancement and flexible prompt learning to explore further and leverage the semantic relationship between vision and language. Rawlekar et al. \cite{102} introduced PositiveCoOp, which learns only positive prompts and replaces negative prompt learning with embedded vectors learned directly in the shared feature space without relying on a text encoder. Liu et al. \cite{103} presented a language-driven cross-modal classifier, which trains a classifier using language data generated by large language models and then transfers it to the visual modality. Fang et al. \cite{104} proposed ProPOT, which generates improved class prototypes by partially aggregating local image features through a partially optimal transport scheme.

Despite these advancements, existing multi-modal learning methods still exhibit limitations. They fail to adequately capture the complex co-occurrence relationships between labels and instances, often overlooking cross-dependencies, which adversely affect alignment performance. Moreover, current models tend to focus on local similarities while lacking the integration of global correlations, thereby constraining the accurate estimation of label confidence. To address these issues, our approach employs global modeling of labels, instances, and their multiple co-occurrence relationships, achieving multi-level semantic fusion and significantly enhancing alignment precision. Furthermore, we construct a cross-modal fusion module to deeply integrate textual and visual data, optimizing label confidence estimation. To address the issue of incomplete label annotations, we propose an intrinsic semantic augmentation strategy, which ensures robust performance.

\begin{figure*}
\centering
\includegraphics[width=1\textwidth]{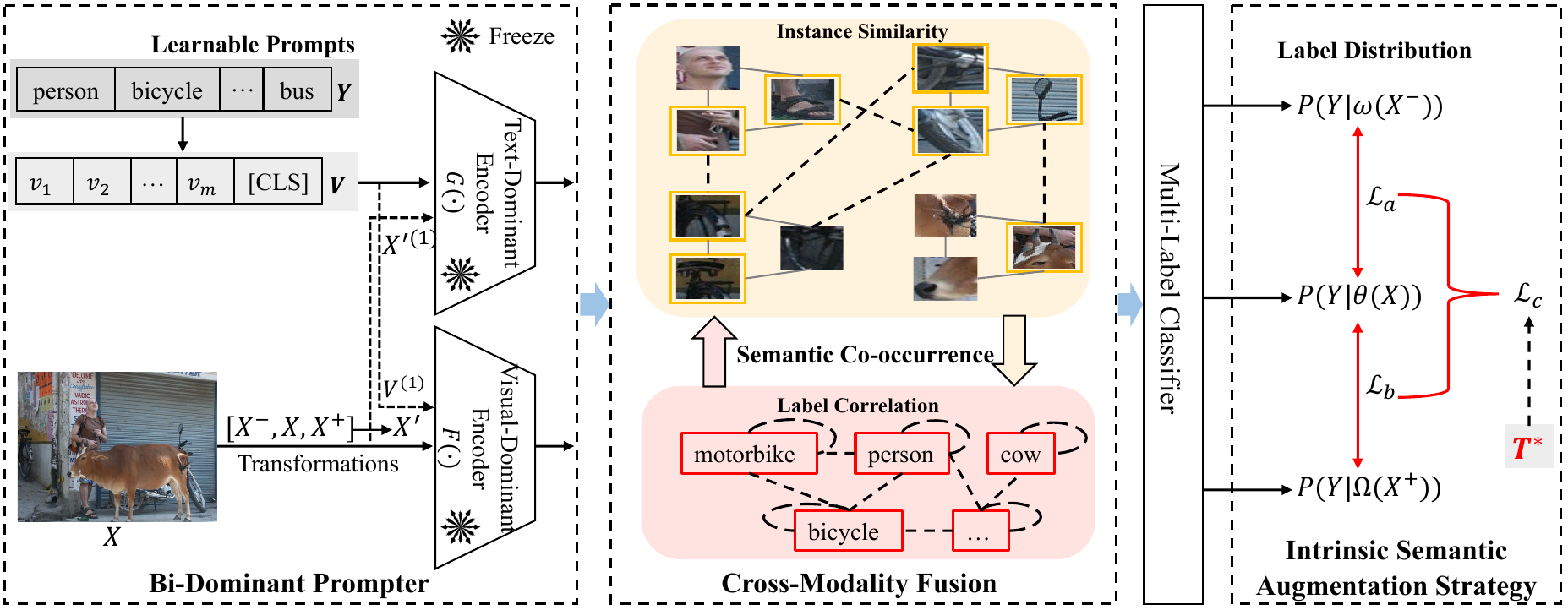}
\caption{Overview of the proposed SCINet method. $X$ consists of instances $s_{i}$, each with known labels $c_{i}$ and unknown labels $u_{i}$. \textcolor{red}{\boldsymbol{$T^{*}$}} denotes the label confidence matrix.} \label{figure3}
\end{figure*}

\section{Methodology}
\label{method}

We propose learning and inferring complex semantic co-occurrence relationships between the visual and textual feature spaces, particularly among labels and labeled instances. This approach enhances the model's ability to identify rare or ambiguous categories. It boosts its generalization skills, enabling it to navigate the complexity and diversity of the real world more effectively. Fig. \ref{figure3} outlines our proposed SCINet method. Initially, we process and interact with input data using the bi-dominant prompter module. Simultaneously, Learnable Prompts acquire textual labels in the form of a vector sequence, providing context for the given label names. Subsequently, text-dominant and image-dominant encoders process textual and image features, respectively. 
The cross-modality fusion module deeply combines text and visual information to understand and classify data more effectively. Furthermore, we utilize an intrinsic semantic augmentation strategy to enable the model to comprehend data features and encourage a synergistic relationship between label confidence and sample complexity.

\subsection{Bi-Dominant Prompter}
Advancements in large-scale pre-trained models, such as BERT and CLIP, facilitate efficient access to contextual label embeddings. This advancement enables the effortless inference of co-occurrence relationships or correspondences between labels and instances. Such a strategy is especially attractive when adequate label supervision is lacking. Furthermore, the extensive prior knowledge contained within pre-trained models facilitates the connection between known and unknown labels, potentially alleviating the challenges associated with limited label supervision. Therefore, we introduce a semantics-driven prompter to explore the label-to-label co-occurrence relationships inherent in the pre-trained model. Given the widespread attention and outstanding performance of vision-language models in computer vision, we have chosen CLIP as the primary focus of our research. Let $V=[v_{1}, v_{2}, \cdots, v_{m}, \text{CLS}]$, where each $v_{i}$ represents a learnable soft prompt token, and CLS denotes the given label name. The entire set of labels is denoted as $Y$, representing each instance as $y_{i} \in Y$. The text-dominant encoder accepts label information while simultaneously processing the image set. The textual representation obtained from the text-dominant encoder process is denoted as $z$. The input image $\mathbf{X}$ is represented as $\mathbf{X}= \left \{ \left (\mathbf{s_{1}},  C_{1}, U_{1}\right ) , \left (\mathbf{s_{2}},  C_{2}, U_{2} \right ) , \cdots, \left (\mathbf{s_{n}},  C_{n}, U_{n} \right ) \right \} $, where $s_{i}$ denotes the $i$-th instance, $C_i = \{\mathbf{c}_{i1}, \mathbf{c}_{i2}, \cdots\}$ represents its set of known correct labels, and $U_i = \{\mathbf{u}_{i1}, \mathbf{u}_{i2}, \cdots \}$ corresponds to its set of unknown labels. To enhance the model's robustness, three different levels of transformations are applied to the input, including weak transformation ($X^{-}$), original transformation ($X$), and strong transformation ($X^{+}$). These three transformations are collectively integrated as $X^{'}$ and utilized for subsequent learning and optimization processes. For a detailed explanation of the three different levels of transformation, please refer to Section \ref{subesection3}. The visual representation obtained, denoted as $f$, is extracted through the modified CLIP image encoder. This dual-encoder approach enables the model to understand and apply the semantic co-occurrence knowledge, thereby improving the performance of MLR. Specifically, the text-dominant encoder and the visual-dominant encoder are defined as follows:
\begin{equation}
 z_{i}=G\left ( V,{X}'^{\left ( 1 \right )}  \right ),
\end{equation}
\begin{equation}
f=F\left ({X}' , V^{\left ( 1 \right )}  \right ).
\end{equation}
Here, $G\left (\cdot \right)$ denotes the text-dominant encoder and $F\left (\cdot \right)$ represents the visual-dominant encoder, both of which are derived from the modified CLIP. To introduce text or image information from different modalities, $V^{\left ( 1 \right )}$ and ${X}'^{\left ( 1 \right )} $ respectively represent the aligned $V$ and ${X}'$.

\subsection{Cross-Modality Fusion Module}

The cross-modality fusion module employs a fusion strategy to optimize label confidence, utilizing both the selected visual features and the optimized label confidence to train the classifier. This module is specifically designed for deep, interactive fusion of information from multiple modalities, such as textual and visual data. It considers the local similarities between samples and global correlations among labels. This fusion strategy enhances the model's performance when handling partial multi-label data, particularly in the presence of noisy labels. In this way, the algorithm can estimate the confidence of each label more accurately, thereby providing the model with a more reliable learning basis.

In the problem of PML, each instance may be associated with a set of candidate labels, some of which may be irrelevant to the problem. We construct an objective function that considers both global instance similarity and local label relevance to compute the confidence of each label.

Assume the distance between instances and a defined domain radius $R$ such that
\begin{equation}
R_{s_{i} } =\left \{ s | s \in X, \bigtriangleup \left ( s, s_{i}  \right )  \le  R \right \},
\end{equation}
where $R$ is the domain radius and $R_{s_{i} }$ denotes the set of all samples within the distance $R$ from instance $s_{i}$.

We evaluate the similarity between instances through the Gaussian function, denoted as $S$, and also account for their proximity relationships to construct a matrix of local sample similarities. To simplify the computation, we have omitted the normalization factor. However, this will be taken into consideration during the subsequent model training phase, thus it will not affect the final performance of the model.
\begin{equation}
S_{ij} =\begin{cases}
 -exp\left ( \left \| s_{i} - s_{j} \right \| _{2}^{2} / 2 \sigma^{2}   \right  ),  & \text{ if } s_{i} \in R_{s_{i}} \; \text{or} \; s_{j} \in R_{s_{i}},\\
 0, &  otherwise.
\end{cases}
\end{equation}

Calculating label confidence is crucial for assessing the reliability of different labels when resolving such issues. Moreover, evaluating the global relevance of labels is also crucial, as it can reveal potential interconnections among them. We employ the Pearson correlation coefficient to calculate the label correlation, denoted as $r$, to quantify the inter-label relevance. This approach enables us to effectively identify and understand the interrelationships among labels, thereby achieving more accurate predictions in multi-label classification tasks. However, it is essential to recognize that this coefficient does not convey information regarding the causal relationships between labels. Even if there is a high correlation between two labels, it does not directly imply that one label causes changes in the other.

The Pearson correlation coefficient $r_{ij}$ is calculated as:
\begin{equation}
r_{ij} =\frac{\sum_{k=1}^{n} \left ( y_{k_{i}} - \bar{y_{i}}  \right ) \left ( y_{k_{j}} - \bar{y_{j}}  \right )}{\sqrt{\sum_{k=1}^{n} { \left (  y_{k_{i}} - \bar{y_{i}}  \right )^{2}\left ( y_{k_{j}} - \bar{y_{j}} \right ) ^{2}  } } } ,
\end{equation}
where $y_{k_{i}}$ and $y_{k_{j}} $ represent the values of the ${K}^{th}$ sample for labels $i$ and $j$ respectively, $\bar{y_{i}}$ and $\bar{y_{j}}$ represent the mean values of labels $i$ and $j$, and $n$ denotes the number of samples. A $r_{ij}$ greater than 0 indicates a positive correlation between labels $i$ and $j$; a $r_{ij}$ equal to 0 indicates no correlation; and a $r_{ij}$ less than 0 indicates a negative correlation between labels $i$ and $j$.

By integrating the instance similarity formula and the label correlation formula, we consider the interactions between sample features and labels when calculating label confidence, thereby improving the reliability of the confidence computation. This semantic co-occurrence not only captures the latent dependencies between labels but also leverages similarities across instances, further enhancing the accuracy of label prediction. By balancing both label and instance-level information, the model gains a more comprehensive understanding of diverse features in complex scenarios, improving its generalization and robustness. Hence, the final confidence matrix is formulated as follows:

\begin{equation}
\begin{aligned}
T^{*} = & \min_{T} \left \| T-Y \right \| _{F}^{2} \\
& +\lambda _{n} \min_{T} \sum_{i=1}^{n} \sum_{j=1}^{n}S_{ij} \left \| T_{i} -T_{j} \right \| ^{2} \\
& +\lambda _{q} \min_{T} \sum_{i=1}^{q} \sum_{j=1}^{q}r_{ij} \left \| T_{i} -T_{j} \right \| ^{2},
\end{aligned}
\end{equation}
where $\left \| \cdot  \right \| _{F}^{2}$ denotes the Frobenius norm, $\left \| \cdot  \right \| ^{2}$ represents the spectral norm, $T$ is the computed label confidence. Given that both known correct and incorrect labels are used to construct the label confidence matrix, it helps infer relationships among labels. Unknown labels are estimated using semantic co-occurrence and cross-modality fusion methods. $ \lambda _{n} $ and $ \lambda _{q} $ are the trade-off parameters for samples and labels, respectively.

\subsection{Intrinsic Semantic Augmentation Strategy}
\label{subesection3}
The proposed intrinsic semantic augmentation strategy aims to boost model performance in situations where labeling is incomplete. By deeply understanding and leveraging semantic information, particularly the co-occurrence relationships between labels, the Intrinsic semantic augmentation strategy enables the model to capture the inherent connections between labels more effectively. This approach acknowledges that models can achieve better results in MLR tasks by fully exploiting these underlying semantic associations, even when faced with partial or missing labels.

In multi-label classification, each instance may belong to multiple categories. Existing loss functions typically reweight each label independently, without considering the inter-label and inter-instance correspondences. We propose a triple transformation strategy to enhance the model's understanding of images to address this issue. By subjecting the input image to weak, medium (i.e., the untransformed original image), and strong transformations, we strive for the model to learn richer information from disturbances of varying degrees. During each augmentation, weak and strong transformations are applied by randomly selecting one out of one and five types of transformations, respectively. These three transformations correspond to different semantic distributions: $P\left ( Y | \omega \left ( X^{-} \right )  \right ) $, $P\left ( Y | \theta \left ( X \right )  \right ) $, and $P\left ( Y | \Omega  \left ( X^{+} \right )  \right ) $, all in $\mathbb{R} ^{n\times 1} $. The weak transformation $\omega \left ( X^{-} \right )$ is designed to make subtle adjustments to the image, such as random cropping, horizontal flipping, and color jittering, preserving most of the original semantic information, thereby aiding the model in identifying and learning the core elements within the image. The medium transformation $\theta \left ( X \right )$ provides a robust baseline, ensuring that the model can accurately capture the fundamental features of the image without being disrupted by the transformation. On the other hand, the strong transformation $\Omega  \left ( X^{+} \right )$ modifies the image in a more aggressive manner, such as random rotation, mixup, and cutmix, introducing greater sample diversity and promoting the model’s ability to extract features in a more flexible and robust way. Subsequently, we construct a confident label set $\mathcal{C} \left ( x \right )$ to enhance consistency among the three transformations. For each label category, a dynamic thresholding strategy is implemented: A label $c$ is only incorporated into the consistency loss computation when its probability surpasses the predefined threshold $\mathcal{K}$. This mechanism effectively mitigates interference from uncertain labels while ensuring consistent optimization of identical labels across multiple transformations. It is worth noting that although the weak transformation $\omega \left ( X^{-} \right )$  and the strong transformation $\Omega  \left ( X^{+} \right)$ perturb the images to different extents, both are processed by the same visual encoder. This ensures that the features are mapped to a unified space, thereby facilitating contrastive learning and consistency constraints. Sharing the encoder also promotes parameter sharing and regularization, enabling the model to focus on capturing the core semantics of the images and thereby enhancing its robustness and generalization capabilities.

The consistency loss between weak transformation $\omega \left ( \cdot \right )$ and medium transformation $\theta \left ( \cdot \right )$, abbreviated as $\mathcal{L} _{a} $:
\begin{equation}
\begin{aligned}
\mathcal{L} _{a} = & -\sum_{c\in \mathcal{C} \left ( x \right ) }\left (  \log{\left ( p|\omega \left ( X^{-} \right ) \right )  } + \log\left ( {p|\theta  \left ( X \right )} \right )  \right ) \\
& -\sum_{c\notin \mathcal{C} \left ( x \right )}\left (  \log{\left (1- p|\omega \left ( X^{-} \right ) \right )  } + \log\left ( {1-p|\theta  \left ( X \right )} \right )  \right ).
\end{aligned}
\end{equation}

The consistency loss between medium transformation $\theta \left ( \cdot \right )$ and strong transformation $\Omega \left ( \cdot \right )$, abbreviated as $\mathcal{L} _{b} $:

\begin{equation}
\begin{aligned}
\mathcal{L} _{b } = & -\sum_{c\in \mathcal{C} \left ( x \right )}\left (  \log{\left ( p|\theta \left ( X \right ) \right )  } + \log\left ( {p|\Omega  \left ( X^{+} \right )} \right )  \right ) \\
& -\sum_{c\notin \mathcal{C} \left ( x \right )}\left (  \log{\left (1- p|\theta \left ( X \right ) \right )  } + \log\left ( {1-p|\Omega  \left ( X^{+} \right )} \right )  \right ).
\end{aligned}
\end{equation}

In practice, data transformations may not yield consistent outcomes. This implies that for learning algorithms, the introduction of adaptability strategies and semantic structure enhancements can help the model better understand and deal with inconsistencies in data structures. Considering the adoption of three unique transformation mechanisms, we perform internal knowledge transfer through a self-distillation strategy to enhance the model's efficacy, leveraging knowledge accumulated at different stages of the training process. By optimizing the Kullback-Leibler divergence, we guide the semantic distribution calibration of transformed images to improve their distribution:
\begin{equation}
\begin{aligned}
\mathcal{L}_{c} &= -\sum_{c\in \mathcal{C} \left ( x \right )} \left( q_c^{\omega} \log\frac{q_c^{\theta}}{q_c^{\omega}} + (1 - q_c^{\omega}) \log\frac{1 - q_c^{\theta}}{1 - q_c^{\omega}} \right) \\
&\quad -\sum_{c\notin \mathcal{C} \left ( x \right )} \left( q_c^{\Omega} \log\frac{q_c^{\theta}}{q_c^{\Omega}} + (1 - q_c^{\Omega}) \log\frac{1 - q_c^{\theta}}{1 - q_c^{\Omega}} \right),
\end{aligned}
\end{equation}
where $q_c^{\omega} = P\left(c|\omega \left(X^{-}\right)\right)$, $ q_c^{\theta} = T^{*} P\left(c|\theta \left(X\right)\right)$, and $q_c^{\Omega} = P\left(c|\Omega \left(X^{+}\right) \right)$.

These three transformations complement each other to form a comprehensive learning strategy that enhances the model's overall understanding and adaptability to data. Subtle adjustments in the weak transformations enable the model to identify and learn the core elements of the image. The medium transformations provide a reliable baseline, ensuring the model can accurately capture the essential features of the image without any interference from transformations. On the other hand, vigorous transformations introduce sample diversity through larger image modifications, encouraging more robust and flexible feature extraction capabilities.

Pareto optimality emphasizes the trade-offs and balances in multi-objective optimization to avoid weakening other objectives while improving one \cite{69,39}. Inspired by this, we combine three different losses and utilize Pareto optimization theory to balance multiple loss functions. We compile the multiple losses into a list, calculate the Pareto frontier, and update the loss weights accordingly.

\section{Experiments}
\label{experiment}

\subsection{General Settings}

\begin{table*}[h]
\caption{Comparison of mAP (\%) with the single label MLR methods.}
\label{table1}
\centering
\begin{tabular}{ccccc|cccccc}
\toprule
\raisebox{-1.5ex}[0pt][0pt]{Method} & \multicolumn{4}{c|}{LargeLoss setup \cite{13}}  & \multicolumn{4}{c}{SPLC setup \cite{14} } \\

 & VOC2012 & COCO2014 & CUB & Avg. mAP & VOC2012 & COCO2014 & CUB & Avg. mAP \\ 
\midrule
LSAN \cite{29} & 85.19 & 64.25 & 14.32 & 54.59 & 85.13 & 64.75 & 15.21 & 55.03 \\
ROLE \cite{29}& 86.20 & 66.27 & 16.30 & 56.26 & 86.19 & 66.64 & 16.92 & 56.59 \\
LargeLoss \cite{13} & 84.46 & 71.07 & 16.45 & 57.33 & 87.52 & 69.88 & 16.75 & 58.05 \\
Hill  \cite{14}& 85.75 & 73.54 & 19.04 & 59.44 & 87.98 & 72.36 & 16.37 & 58.90 \\
SPLC  \cite{14} & 86.53 & 73.45 & 19.30 & 59.76 & 88.50 & 72.05 & 18.07 & 59.54 \\
BoostLU \cite{30} & 88.69 & 73.64 & 19.26 & 60.53 & 87.77 & 72.35 & 17.47 & 59.20 \\
SCPNet \cite{8}& 90.52 & 75.43 & 23.55 & 63.17 & 90.55 & 75.92 & 24.28 & 63.58 \\ 
 SCINet (ours) & \textbf{90.97} & \textbf{75.52} & \textbf{26.16} & \textbf{64.21} & \textbf{91.76} & \textbf{76.46} & \textbf{26.16} & \textbf{64.79} \\ 
\bottomrule
\end{tabular}
\end{table*}

\subsubsection{Datasets} We conducted extensive experiments on several standard PML benchmarks for incomplete label settings. For the single positive label setting, as referenced in \cite{8, 13, 14}, we utilized VOC2012 \cite{27}, COCO2014 \cite{25}, and CUB \cite{26}. For partial label learning, we employed VOC2007 \cite{28} and COCO2014 \cite{25}.

\subsubsection{Implementation details} To ensure a fair comparison with other methods, we conducted experiments with the CLIP method based on the ResNet50 architecture for both single positive and partial labels. We utilized an Adam optimizer with a learning rate of 4e-5 and a weight decay of 1e-4. The training was carried out on an RTX A5000. The full implementation and additional details are available in the repository\footnote{ \textcolor{magenta}{\url{https://wuliwuxin.github.io/SCINetProject/}}}.

\subsubsection{Baseline Methods} To validate the proposed SCINet approach, several state-of-the-art methods are selected for comparison in the experimental evaluation.
\paragraph{Conventional MLR Algorithms}
\textbf{SSGRT} \cite{16} proposes a method that incorporates semantic decoupling and interaction modules.
\paragraph{Weakly Supervised MLR Algorithms}
\textbf{LargeLoss} \cite{13} introduces a method for weakly supervised multi-label classification by modifying large-loss samples to prevent memorization of noisy labels.
\textbf{LSAN} \cite{29} and \textbf{ROLE} \cite{29} propose effective strategies for multi-label learning from single positive labels through tailored loss designs.
\textbf{Hill} \cite{14} and \textbf{SPLC} \cite{14} present loss functions designed to handle multi-label learning with missing labels.
\paragraph{General Algorithms}
\textbf{BoostLU} \cite{30} introduces a method aimed at enhancing attribution scores.
\textbf{SCPNet} \cite{8} explores a structured semantic prior for improved learning.
\textbf{SST} \cite{31} and \textbf{HST} \cite{60} investigate intra-image and cross-image semantic correlations to boost performance.
\textbf{SARB} \cite{32} blends category-specific representations across different images to enhance multi-label learning.
\textbf{DualCoOp} \cite{7} utilizes dual learnable prompts for improved model adaptation.
\textbf{MLR-PPL} \cite{70} features a method with category-adaptive label discovery and noise rejection mechanisms.

%

\subsubsection{Evaluation} To facilitate a fair comparison, we default to using the mean average precision (mAP) in Eq. (\ref{equation11}) as the evaluation metric, consistent with prior studies. For the single positive label setting, we randomly retain one positive label and treat the remaining labels as unknown, adopting two different configurations: the LargeLoss setting and the SPLC setting \cite{13, 14}. We also conduct thorough evaluations in the partial label setting by using both the overall F1 score (OF1) and the per-class F1 score (CF1), as detailed in Eq. (\ref{equation14}) and Eq. (\ref{equation17}). Additionally, we randomly reserve a portion of labels for the training set, with the proportion increasing arithmetically by 20\% from 10\% to 90\%.

\begin{equation}
\label{equation10}
   {\rm A{{P}_{i}}}=\frac{1}{\left| {{G}_{i}} \right|}\underset{k=1}{\overset{n}{\mathop \sum }}\,{{P}_{k}\times re{{l}_{k}}},  \\
\end{equation}

\begin{equation}
\label{equation11}
     {\rm mAP}=\frac{1}{\left| \mathcal{Y} \right|}\underset{i=1}{\overset{\left| \mathcal{Y} \right|}{\mathop \sum }}\,\text{A}{{\text{P}}_{i}},  \\
\end{equation}
where $|{{G}_{i}}|$ represents the number of samples in the $i$-th category, ${{P}_{k}}$ represents the precision of the top $k$ predictions, ${rel}_{k}$ represents whether the $k$-th prediction is the true label of the sample (1 if yes, 0 if no), $n$ represents the total number of predictions, $y$ represents the set of all labels, and ${\rm {A}{{P}_{i}}}$ represents the average precision for the $i$-th label.

\begin{equation}
\label{equation14}
     {\rm OF1}=\frac{2\times TP}{2\times TP+FP+FN},  \\
\end{equation}

\begin{equation}
\label{equation17}
    {\rm  CF1}=\frac{1}{K}\underset{i=1}{\overset{K}{\mathop \sum }}\,\frac{2\times T{{P}_{i}}}{2\times T{{P}_{i}}+F{{P}_{i}}+F{{N}_{i}}},  \\
\end{equation}
where $K$ represents the total number of categories, $TP_{i}$ represents the number of true positives for $i$-th category, $FP_{i}$ represents the number of false positives for $i$-th category, and ${{FN}_{i}}$ represents the number of false negatives for $i$-th category.

\subsection{Comparisons with State-of-the-Arts}

\begin{table*}[ht]
\centering
\caption{Comparison of the proposed SCINet model with the current advanced competitors regarding average mAP, OF1, and CF1 (\%). ‘*’ denotes the performance reported in their paper.}
\label{table2}
\begin{tabular}{cccccccccc}
\toprule
Datasets & Methods & 10\% & 30\% & 50\% & 70\% & 90\% &Avg.mAP & Avg. OF1 & Avg. CF1\\ \midrule
\multirow{6}{*}{VOC2007}   
& SSGRT \cite{16} (ICCV '19)   & 75.77 & 87.01 & 90.36 & 91.30 & 92.42 & 87.37 & 43.88 & 43.73  \\
& SST \cite{31} (AAAI '22)   & 85.51 & 90.44 & 91.99 & 92.21 & 92.78 & 90.59& 87.76 & 85.74 \\ 
& SARB \cite{32} (AAAI '22)    & 84.11 & 90.49 & 91.97 & \textbf{92.54} & 92.68 & 90.36& 87.46 & 84.63 \\ 
& DualCoOp \cite{7}(NeurIPS '22)  & 90.94 & 91.28 & 92.43 & 92.48 & 92.85 & 91.99  & 85.99 & 84.20\\ 
& SCPNet \cite{8} (CVPR '23) & 91.56 & 91.85 & 91.88 & 91.91 & 91.92 & 91.82 & 86.35 & 85.44 \\ 
& MLR-PPL \cite{70}*  (TMM '24)   & 76.30&	86.40&	88.60	&90.00	&91.90	&86.64	&-	&- \\ 
& HST \cite{60}*  (IJCV '24)  & 84.30&	90.50&	91.60	&92.50	&92.80	&90.34	&-	&- \\ 
& SCINet (ours)     & \textbf{92.32} & \textbf{92.39} & \textbf{92.53} & \textbf{92.54} & \textbf{92.87} & \textbf{92.53} & \textbf{87.83} & \textbf{86.22}\\
\midrule
\multirow{5}{*}{COCO2014} 
& SST \cite{31}  (AAAI '22)     & 67.37 & 68.32 & 70.45 & 71.57 & 72.52 & 70.05  & 70.12 & 65.41\\
& SARB \cite{32} (AAAI '22)    & 65.11 & 70.32 & 74.10 & 74.19 & 74.55 & 71.65 & 71.18 & 67.54\\
& DualCoOp \cite{7} (NeurIPS '22) & 75.61 & 77.26 & 77.53 &77.77& 78.81 & 77.40 & 58.55 & 55.68\\
& SCPNet \cite{8} (CVPR '23)  & 75.78 & 75.86 & 75.87 & 75.90 & 75.96 & 75.87 & 71.00 & 64.79\\ 
& MLR-PPL \cite{70}*  (TMM '24) & 40.40&	67.20&	74.10	&77.00	&78.90	&67.52	&-	&- \\ 
& HST \cite{60}* (IJCV '24) & 70.60&	77.30&	\textbf{79.00}	&\textbf{79.90}	&\textbf{80.40}	&77.44	&-	&- \\ 
& SCINet (ours)     & \textbf{77.65} &\textbf{77.69} & 77.71 & 77.73 & 78.86 & \textbf{77.93} & \textbf{71.97} & \textbf{70.86}\\ 
\bottomrule
\end{tabular}
\end{table*}

\begin{figure}
\centering
\includegraphics[width=0.48\textwidth]{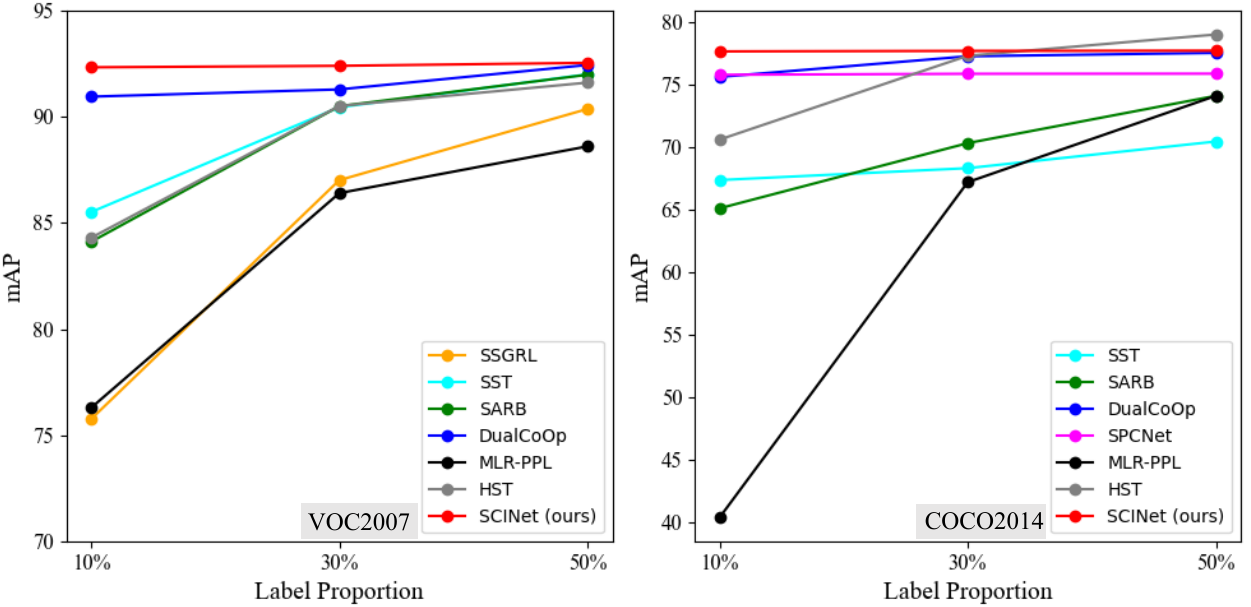}  
\caption{Our SCINet model and the current advanced methods are compared in terms of mAP on the VOC2007 and COCO2014 datasets, with known label proportions set to increase arithmetically by 20\%, ranging from 10\% to 50\%.} 
\label{figure4}
\end{figure}

This section assesses the efficacy of the proposed SCINet in single-label scenarios by comparing it with seven leading methodologies across three datasets. Furthermore, we evaluate SCINet's efficacy on partially labeled data by randomly preserving a subset of labels in two datasets and contrasting it with six cutting-edge methodologies. Table \ref{table1} presents the mAP and average mAP outcomes of several methods on the VOC2012, COCO2014, and CUB datasets. Table \ref{table2} presents the mAP, average mAP, OF1, and CF1 scores across different fractions of incomplete labels for the VOC2007 and COCO2014 datasets. Fig. \ref{figure4} illustrates the trend of partial label performance on the VOC2007 and COCO2014 datasets. Based on these empirical findings, we can derive the subsequent conclusions:
\begin{itemize}
    \item  We performed studies on six cases in the two single-label settings (2 settings × 3 datasets = 6 cases). The findings are presented in Table \ref{table1}. The proposed SCINet attained optimal performance in all six instances, achieving a score of 100.00\%. In the VOC2012 dataset, SCINet attained a mAP of 90.97\% in the LargeLoss configuration and 91.76\% in the SPLC configuration, exceeding the performance of the leading existing models by 0.45\% and 1.21\%, respectively. These findings illustrate SCINet's exceptional efficacy on the VOC2012 dataset, especially in fine-grained classification tasks. Comparable patterns were noted in the other datasets (COCO2014 and CUB), where SCINet surpassed current state-of-the-art techniques across various configurations, with the greatest significant enhancement observed in the CUB dataset.
     \item According to Table \ref{table1}, SCINet's avg. mAP increased by 1.04\% and 1.21\% under the LargeLoss and SPLC settings, respectively. These results indicate that SCINet demonstrates exceptional performance and generalization ability in handling these tasks, particularly in fine-grained classification.
     \item For the partially labeled multi-label experiments, we executed 16 cases (8 metrics × 2 datasets = 16 cases), with the results displayed in Table \ref{table2}. The proposed SCINet achieved the best performance in 13 cases, accounting for 81.25\%. SCINet surpassed prior state-of-the-art models on the VOC2007 dataset across various performance parameters. Meanwhile, SCINet enhanced the avg. mAP by 2.19\% compared to the prior leading state-of-the-art approach (HST). Moreover, using merely 10\% of the training data, SCINet attained an average mAP of 92.32\%, surpassing the HST by 8.02\%. These results illustrate SCINet's substantial enhancement in performance when managing restricted annotation data. In the COCO2014 dataset, SCINet demonstrated a comparable performance enhancement trend to that observed in VOC2007, highlighting its consistency and excellence across diverse datasets.
     \item As demonstrated in Fig. \ref{figure4}, SCINet continuously outperforms as the annotation fraction rises, particularly at lower annotation proportions. The enhancement in performance is more seamless, demonstrating exceptional robustness and generalization capability. While HST exhibits a marginal advantage at the 50\% annotation level on the COCO2014 dataset, SCINet nonetheless displays superior overall stability and consistency.
      \item Fig. \ref{figure4} illustrates that the mAP of all models progressively enhances with the rise in annotation proportion. The mAP of all models has an upward trajectory from 10\% to 50\%, signifying that a greater annotation proportion enhances model performance.
\end{itemize}
In summary, SCINet exhibits exceptional performance across various datasets and contexts, notably excelling in fine-grained classification tasks in situations with restricted label proportions. SCINet routinely surpasses existing state-of-the-art approaches, demonstrating its strong generalization abilities and notable performance improvements in multi-label classification tasks, regardless of the quantity of training data used, whether modest or extensive.

\begin{table*}[]
\centering
\caption{The impact of different modules within the SCINet method on results for single positive and partial label settings (\%).}
\label{table3}
\begin{tabular}{c|cc|ccc|ccc|cc|c}
\toprule
\multirow{2}{*}{Method} &\multicolumn{1}{c}{\multirow{2}{*}{BDP}} & \multicolumn{1}{c|}{\multirow{2}{*}{CFM}} & \multicolumn{3}{c|}{SDEL} & \multicolumn{3}{c|}{Single Label}  & \multicolumn{2}{c|}{Partial Label} & \multirow{2}{*}{Avg. mAP} \\ 
                        &  \multicolumn{1}{c}{}  &  \multicolumn{1}{c|}{}  & $\mathcal{L}_{a}$ & $\mathcal{L}_{b}$ & $\mathcal{L}_{c}$ & VOC2012 & COCO2014 & CUB & VOC2007 & COCO2014 & \\ \midrule
Baseline                &   \ding{55}   & \ding{55}     &      \ding{55}                     &            \ding{55}                   &       \ding{55}               & 83.88   & 71.60    & 18.85 & 84.95  & 73.78    & 66.61 \\ \midrule
\multirow{5}{*}{SCINet (ours)} & \checkmark &    \ding{55}  &\ding{55}   &     \ding{55} &  \ding{55}    & 89.39   & 74.72    & 21.14 & 90.37  & 75.40    & 70.20 \\ 
			&\ding{55}  & \checkmark &  \ding{55} &   \ding{55}    &   \ding{55}   &89.50  & 74.15 & 21.70  & 91.74 & 75.47 & 70.51   \\
			&  \ding{55} &   \ding{55}   &   \checkmark  & \checkmark   & \checkmark       &89.58 & 75.42 & 21.68 & 91.87 & 75.55 & 70.82  \\
                        & \checkmark & \checkmark &    \ding{55}     &  \ding{55}  &              \ding{55}        & 90.49   & 75.41    & 22.02 & 92.00  & 77.62    & 71.51 \\
                        & \checkmark & \checkmark & \checkmark  &   \ding{55}  & \ding{55} & 90.93   & 75.50    & 24.32 & 92.08  & 77.63    & 72.10 \\
                        & \checkmark&  \checkmark  &  \ding{55}   & \checkmark   & \ding{55}       &90.74 & 75.57 & 24.19 & 92.00 & 77.65 & 72.03   \\
			& \checkmark &  \checkmark   &  \ding{55}  & \ding{55}  & \checkmark       &  91.07 & 75.97 & 24.45 & 92.12 & 77.65 & 72.25  \\
                        & \checkmark & \checkmark & \checkmark    & \checkmark    &  \ding{55}  & 90.99   & 75.82    & 24.60 & 92.15  & 77.70    & 72.25 \\
                        & \checkmark & \checkmark & \checkmark & \checkmark  & \checkmark          & \textbf{91.76}   & \textbf{76.46}    & \textbf{26.16} & \textbf{93.97}  & \textbf{77.98}    & \textbf{73.27} \\ \bottomrule
\end{tabular}
\end{table*}

\begin{figure}
\centering
\includegraphics[width=0.4\textwidth]{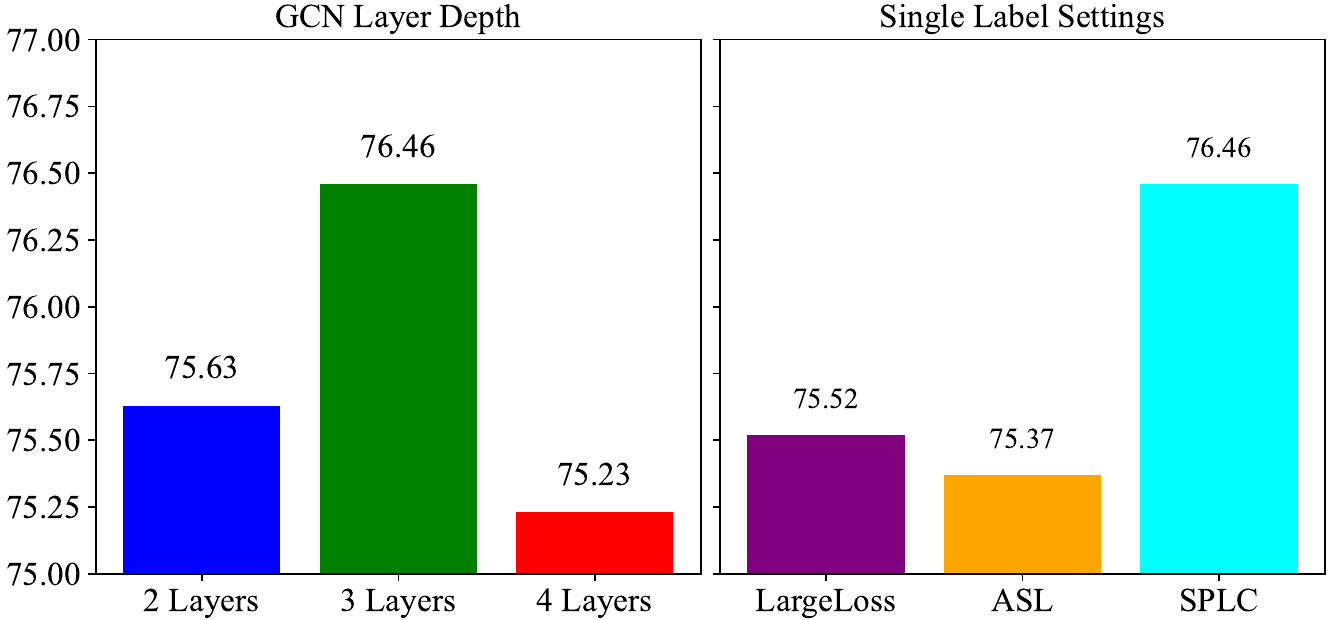}  
\caption{Performance comparison of GCN layer depth and single label settings on the COCO2014 dataset.} 
\label{figure2}
\end{figure}

\subsection{Ablation Study}

The ablation study findings of the SCINet model, as presented in Table \ref{table3}, indicate that the incorporation of each module and approach is crucial for enhancing the model's performance. Through the incremental incorporation of the bi-dominant prompter module, cross-modality fusion module, and an intrinsic semantic augmentation strategy, SCINet markedly enhanced the avg. mAP across various datasets, demonstrating its robust feature extraction and representation learning abilities in multi-label classification tasks. Furthermore, SCINet demonstrated outstanding efficacy in managing label correlation and class differentiation, greatly above baseline models. In instances including semantic co-occurrence, SCINet adeptly differentiated co-existing labels by employing more accurate feature separation and clustering, thus improving the model's adaptability and classification performance in intricate contexts. SCINet exhibits exceptional performance and extensive applicability in complicated tasks, diverse labeling scenarios, and semantic co-occurrence processing, owing to its modular design and strategic tuning.

\textit{Quantitative analysis:} The diverse modules of SCINet are essential for improving the model's performance. The bi-dominant prompter module markedly enhances the model's flexibility to varied data by improving the contextual comprehension of labels, resulting in a 3.59\% improvement in avg. mAP. The introduction of the cross-modality fusion module enhances the model's capacity to differentiate between categories, yielding a 3.90\% improvement in avg. mAP. The integration of several loss functions in the intrinsic semantic augmentation learning strategy demonstrates significant advantages, resulting in a 1.76\% enhancement in avg. mAP. Following the implementation of the bi-dominant prompter and cross-modality fusion modules, the incorporation of $\mathcal{L}_{a}$ alone enhanced the avg. mAP by 0.59\%, whereas the addition of $\mathcal{L}_{b}$ alone yielded a 0.52\% improvement in avg. mAP. Ultimately, the exclusive integration of the $\mathcal{L}_{c}$ loss function resulted in the avg. mAP enhancement of 0.74\%. The findings demonstrate that the modular design and strategic optimization of SCINet substantially enhance the model's performance, enabling it to excel in multi-label classification tasks and underscoring its extensive application potential.

\begin{figure*}
\centering
\includegraphics[width=0.85\textwidth]{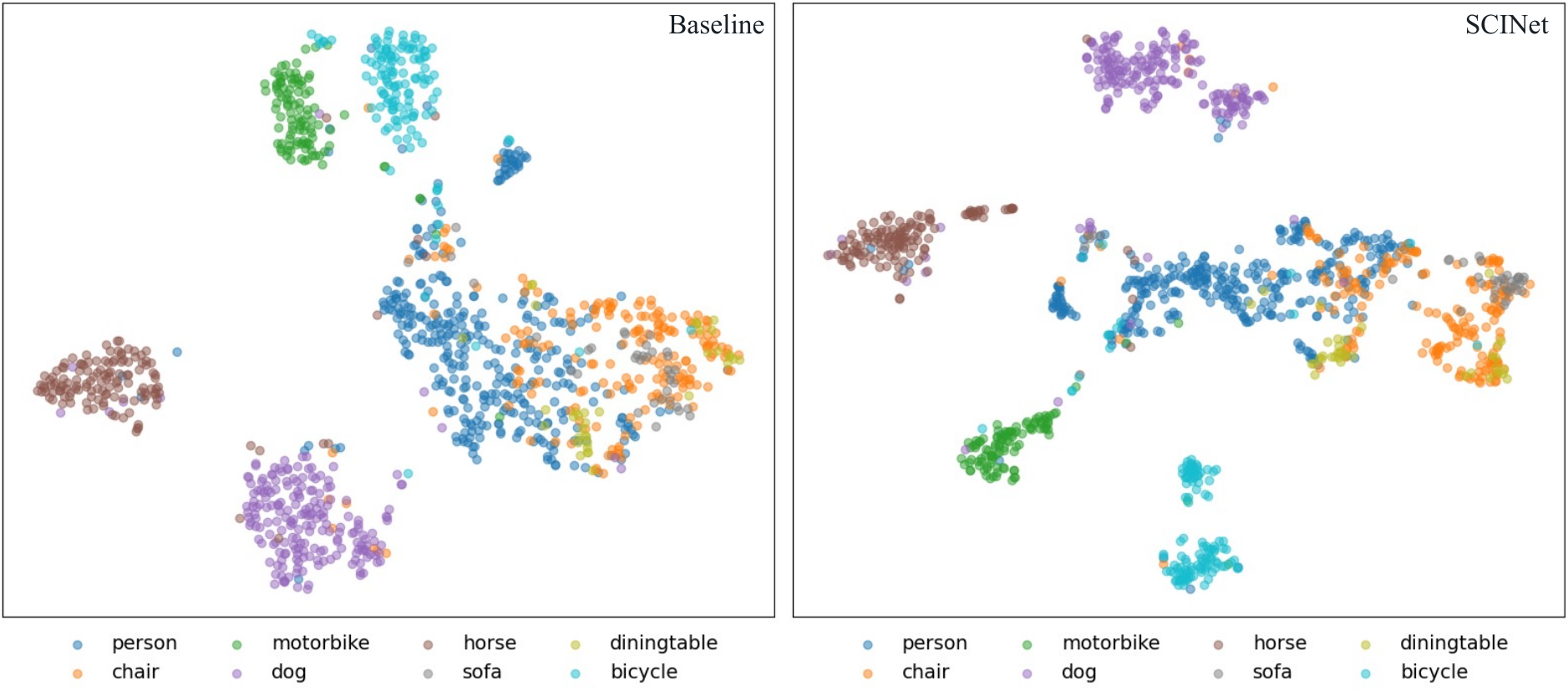} 
\caption{The t-SNE visualization of specific categories, including "person," "chair," and "motorcycle," from the VOC2007 dataset for both the Baseline and SCINet models.} \label{figure5}
\end{figure*}

\begin{figure*}
\centering
\includegraphics[width=1\textwidth]{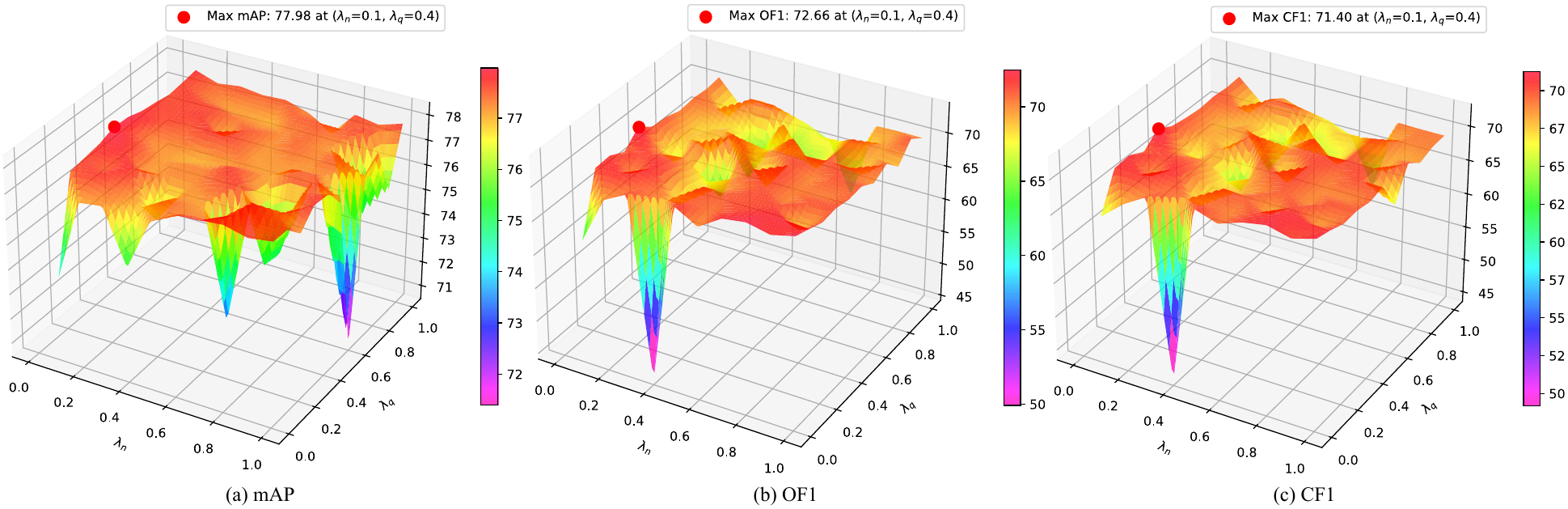}
\caption{The performance metrics of mAP, OF1 and CF1 for trade-off parameters $\lambda_{n}$ and $\lambda_{q}$ vary across the COCO2014 dataset. The optimal values for each metric are marked with red dots within their respective plots.} 
\label{figure6}
\end{figure*}

\begin{figure*}
\label{figure7}
\centering
\includegraphics[scale=0.45]{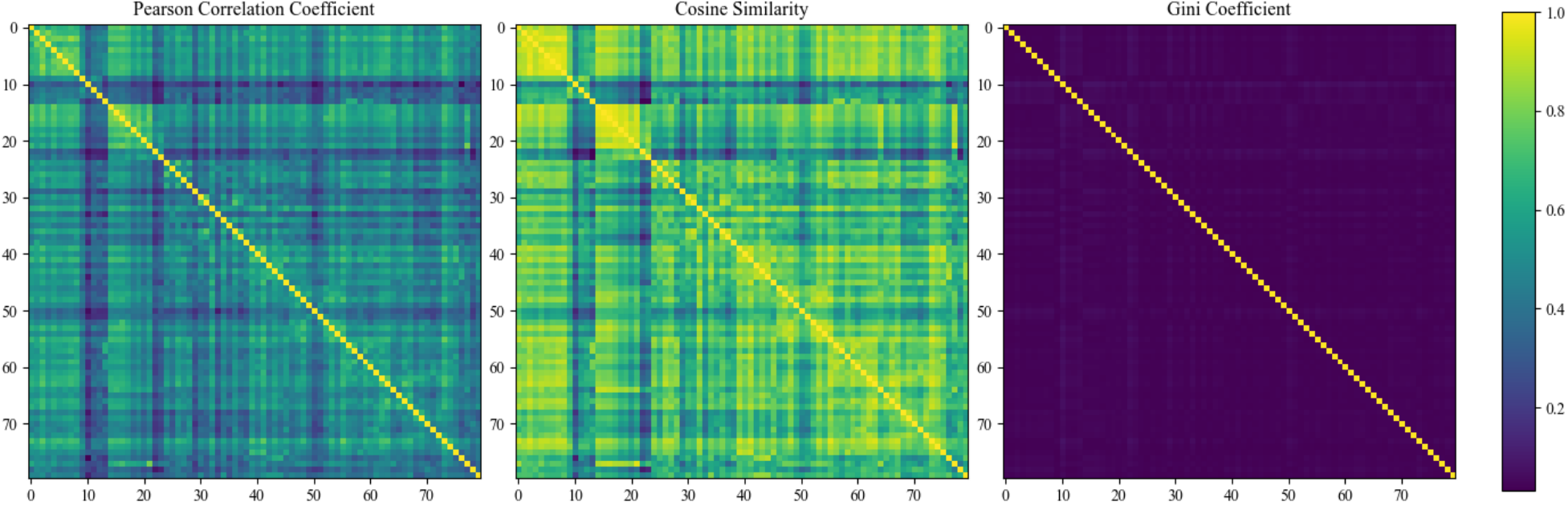} 
\caption{Comparison of different similarity measures on the COCO2014 dataset: Pearson correlation coefficient, Cosine similarity, and Gini coefficient.}
\end{figure*}

%
%

\begin{table}[h]
\centering
\caption{Analysis of label correlation methods on the COCO2014 dataset (\%).}
\label{table7}
\begin{tabular}{cccc}
\toprule
Methods &  mAP & OF1 & CF1  \\ \midrule
Cosine Similarity & 74.30 & 64.17 & 55.84 \\
Gini Coefficient & 73.97 & 64.18 & 55.86 \\
Pearson Correlation Coefficient &  \textbf{77.98} & \textbf{72.66} & \textbf{71.40}  \\ \bottomrule
\end{tabular}
\end{table}

The proposed SCINet exhibits substantial enhancements in dataset performance across various datasets. SCINet attains a 7.88\% enhancement in mAP relative to the baseline model on the VOC2012 dataset. In the COCO2014 dataset with singular labels, the mAP enhancement is 4.86\%, whereas in the CUB dataset, the mAP improvement attains 7.31\%, underscoring SCINet's distinct superiority in fine-grained classification tasks. Furthermore, on the VOC2007 dataset with incomplete labels, SCINet's avg. mAP rises by 9.02\%, whereas on the COCO2014 dataset with incomplete labels, the mAP increase is 4.20\%. The results demonstrate that SCINet can substantially improve model classification performance across diverse label contexts. SCINet significantly outperforms the baseline model across many datasets due to the thorough integration of components, including the bi-dominant prompter module, cross-modality fusion module, and intrinsic semantic augmentation strategy. SCINet's avg. mAP across all datasets exhibits a 6.66\% enhancement relative to the baseline model, underscoring its robust competitiveness and extensive adaptability in multi-label classification tasks. This further substantiates SCINet's capacity to sustain outstanding performance throughout varied and intricate tasks, as well as its considerable advantage in numerous labeling contexts. Furthermore, this study examined the layer depth findings of graph convolutional networks and studied the effects of single-label configurations and label correlation methods on performance, as seen in  Fig. \ref{figure2} and Table \ref{table7}.

\begin{figure*}
\centering
\includegraphics[width=0.8\textwidth]{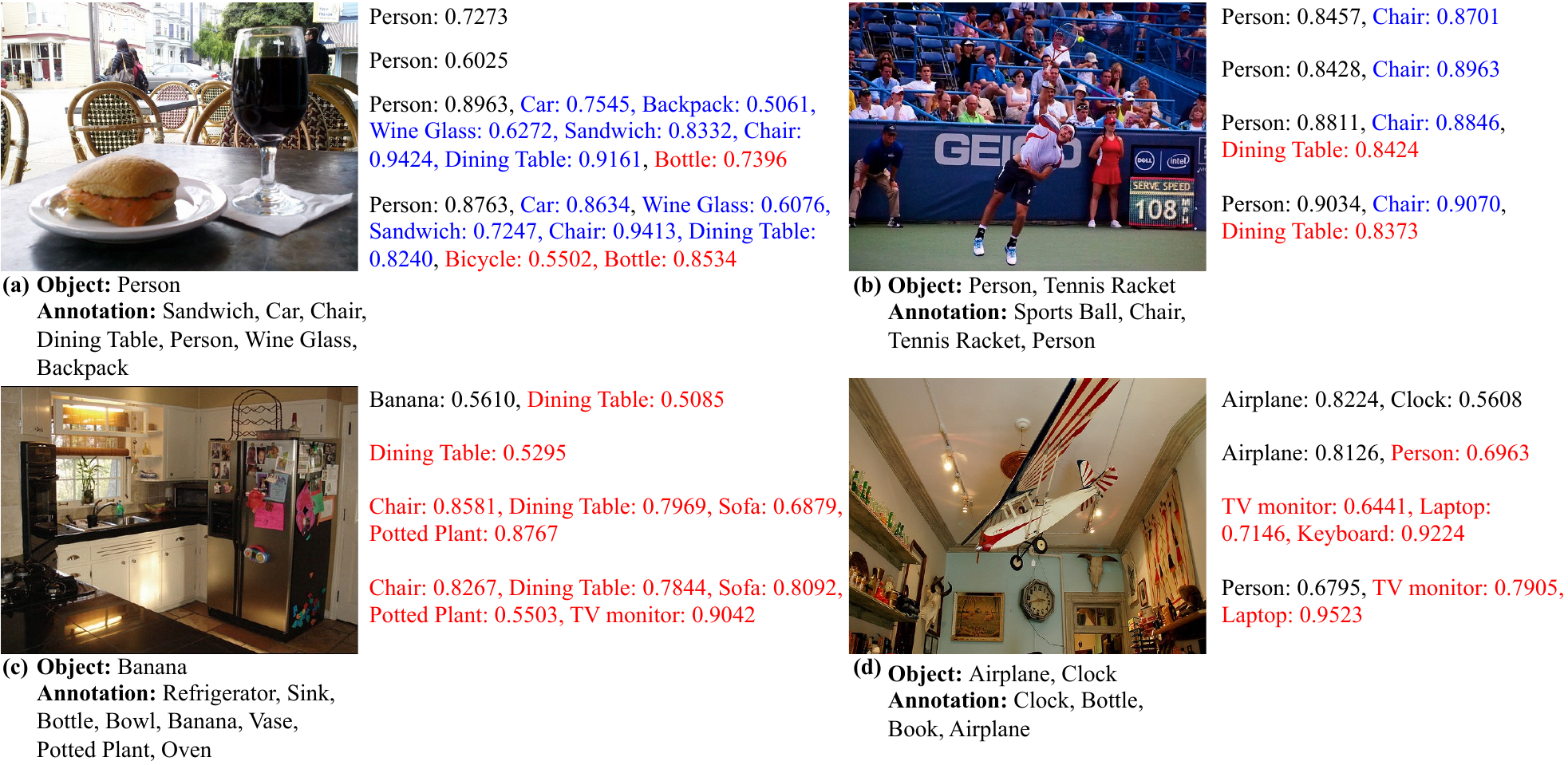}  
\caption{Sample illustrations of different numbers of learnable prompts. Each row in the subfigures corresponds to a different prompt length, ordered from top to bottom as 4, 8, 16, and 32. The \textcolor{black}{black} represents correct predictions, the \textcolor{red}{red} denotes incorrect detection results, and the \textcolor{blue}{blue} indicates detected objects that exist in the label universe but are not part of the target label set.} 
\label{figure9}
\end{figure*}

\begin{figure*}
\centering
\includegraphics[width=0.7\textwidth]{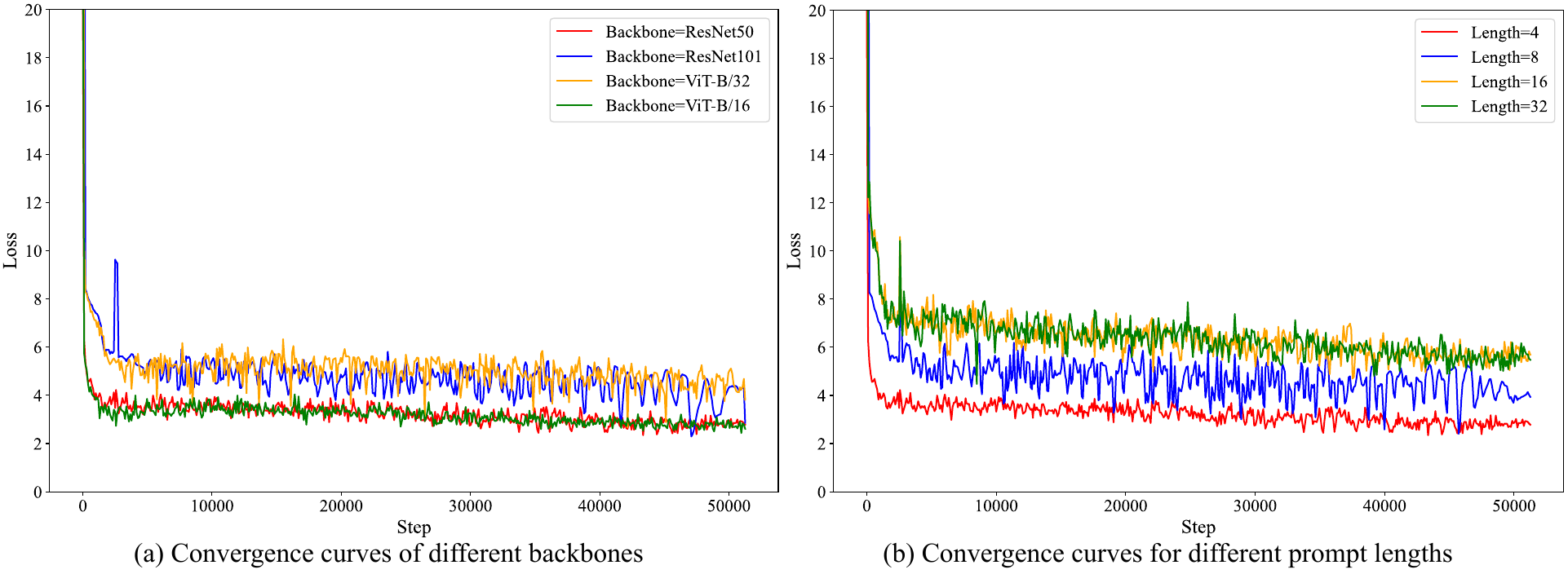}  
\caption{Comparison of the convergence curves across different experimental conditions.} 
\label{figure10}
\end{figure*}

\begin{figure}
\centering
\includegraphics[width=0.33\textwidth]{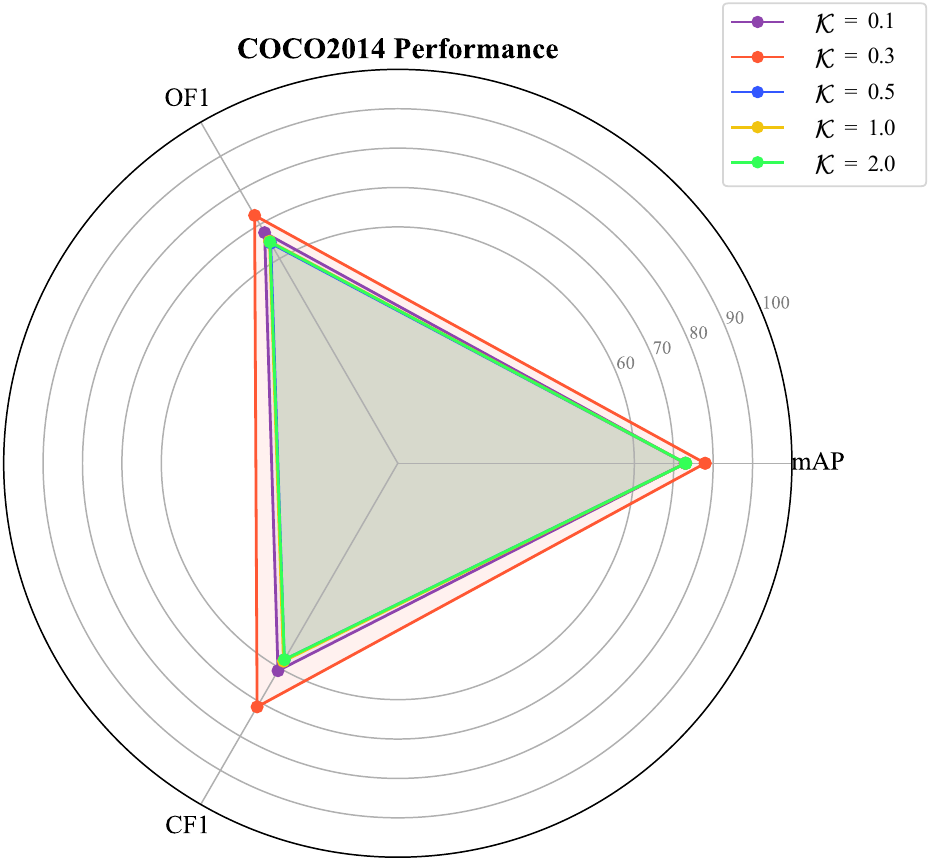}  
\caption{Comparative analysis of different $\mathcal{K}$ on the performance of mAP, CF1, and OF1 metrics for the COCO2014 datasets.} 
\label{figure8}
\end{figure}

\textit{Qualitative analysis:} As illustrated in Fig. \ref{figure5}, SCINet (right Fig. \ref{figure5}) substantially surpasses the Baseline model (left Fig. \ref{figure5}) in managing label co-occurrence. In situations where people and cyclists share the same space, the Baseline model demonstrates significant overlap of feature points for these two categories. Conversely, SCINet demonstrates a clearer differentiation of feature points for these labels, suggesting that SCINet can more precisely differentiate between coexisting pedestrians and bicycles. SCINet proficiently distinguishes data points across several categories, leading to more compact clustering of analogous data points, indicating enhanced abilities in feature extraction and representation learning. This probably results in enhanced classification accuracy and improved generalization. In contrast, the Baseline model exhibits inferior inter-class separation, characterized by considerable overlap among data points from distinct categories and a more dispersed intra-class distribution, signifying diminished class separability. From the standpoint of label correlation, SCINet (right Fig. \ref{figure5}) markedly enhances the distinction between various labels, hence more accurately representing label correlations within the feature space. SCINet distinctly groups highly linked labels while efficiently segregating less correlated labels. SCINet exhibits considerable benefits in label correlation and class separation capabilities. The capacity to more efficiently capture and express label correlations during feature extraction and representation learning boosts inter-class separation and intra-class compactness, hence increasing the model's classification accuracy and generalization abilities. According to the experimental findings, we established $\lambda_{n} = 0.1$ and $\lambda_{q} = 0.4$, as illustrated in Fig. \ref{figure6}. The efficacy of various measurement methods presents distinct advantages and disadvantages, as illustrated in Fig. \ref{figure7}. 

\subsection{Adaptive Dynamic Threshold Analysis}

We evaluated the model's adaptive dynamic threshold, as shown in Fig. \ref{figure8}. By adjusting the threshold parameter $\mathcal{K}$, we observed its impact on the model's attention distribution and performance for larger and more complex datasets, such as COCO 2014. Moderately lowering the temperature value improved the model's performance on key metrics. Notably, when $\mathcal{K}=0.3$, the model demonstrated the best balance between precision and recall, suggesting that lower threshold values can effectively enhance the model's ability to recognize minority classes, which is particularly crucial when dealing with complex datasets. This indicates that dynamically adjusting the threshold $\mathcal{K} $ plays a significant role in optimizing performance on large-scale, complex datasets.

\subsection{Visualization of the Quantity of Learnable Prompts}

Fig. \ref{figure9} illustrates the effect of the number of learnable prompts on object recognition performance across various scenes. As the prompt length increases, the model’s false detection rate (highlighted in red) also increases. For example, in Fig. \ref{figure9}(a), when the prompt length is 4 (first row) and 8 (second row), the detection is entirely accurate. However, when the prompt length increases to 32 (fourth row), false detections become significantly more frequent, with multiple incorrect annotations such as “Bicycle: 0.5502” and “Bottle: 0.8534.” This suggests that, particularly in complex scenes, while increasing prompt length can enhance the model’s detection capabilities, it may also introduce more errors. A similar trend is observed in Figs. \ref{figure9}(b) and \ref{figure9}(d).In Fig. \ref{figure9}(c), when the prompt length is 4 (first row), the model successfully detects the correct object label (“Banana: 0.5610”) but also produces false labels, such as “Dining Table: 0.5085.” This indicates that, with shorter prompts, the model can correctly identify some objects while still suffering from false detections. However, when the prompt length increases to 8, 16, and 32 (second to fourth rows), the detection results exhibit notable errors, including high-confidence false labels like “Chair: 0.8581,” “TV monitor: 0.9042,” and “Dining Table: 0.7844.” These results suggest that in complex scenes containing numerous small objects, shorter prompt lengths help maintain the model’s object recognition capabilitie. In contrast, longer prompt lengths may incorporate additional features that lead to a significant increase in the false detection rate.

\begin{table}[h!]
\caption{Backbone Model Performance Comparison}
\centering
\begin{tabular}{cccccc}
\toprule
Backbone & mAP & OF1 & CF1& Inference Time \\ \midrule
ResNet50 & 77.98 &  \textbf{72.66} &  \textbf{71.40} &  \textbf{623.677ms} \\
ResNet101 & 74.86 & 64.66 & 56.69 & 852.028ms \\
ViT-B/32 & 72.96 & 70.77 & 66.00 & 701.364ms \\
ViT-B/16 & \textbf{82.07} & 70.85 & 71.05 & 1386.979ms \\ \bottomrule
\label{table5}
\end{tabular}
\end{table}

\subsection{Convergence Analysis}

To validate the model's convergence, we plotted the iterative variation curves of the objective function values under different backbone configurations (ResNet50, ResNet101, ViT-B/32, and ViT-B/16) and various context lengths (4, 8, 16, and 32), as illustrated in Fig. \ref{figure10}. Overall, as the number of iterations increases, the loss value decreases rapidly and gradually stabilizes, ultimately reaching a satisfactory steady state, which substantiates the model’s robust convergence properties. Furthermore, the combination of different backbones and context lengths exhibits variations in convergence speed. Table \ref{table5} presents a performance comparison of different backbone networks across multiple metrics.

\subsection{Limitation and Future Work}
While SCINet demonstrates state-of-the-art performance in PML tasks, it still presents certain limitations. Visual analysis reveals that increasing the length of learnable prompts can enhance detection capabilities; however, this often comes at the cost of a higher false detection rate, particularly in complex scenes with numerous small objects. This suggests a trade-off. Longer prompts may introduce additional, potentially noisy features, thereby undermining the model’s robustness and generalization ability, especially in scenarios characterized by occlusions or background clutter.

A key aspect of future work involves conducting more fine-grained and interpretable analyses, such as label-specific detection and language decomposition, to uncover the intricate internal structures and semantic nuances associated with each label. This includes investigating adaptive prompt learning strategies that dynamically adjust prompt length in response to scene complexity, thereby maintaining recognition accuracy while minimizing erroneous detections. Furthermore, the coming research will focus on optimizing the model architecture to improve adaptability and robustness across a broader range of challenging conditions. 

\section{Conclusions}
\label{conclude}
In this study, we introduce SCINet, a novel framework designed to tackle the challenges of PML tasks. At the core of SCINet is the use of a structured semantic prior from multi-modal representation learning to compensate for the lack of supervisory information by introducing a semantic co-occurrence insight network. SCINet combines text and visual features through a cross-modal fusion module, enhancing the model's contextual understanding, feature extraction efficiency, and classification accuracy. Moreover, through the intrinsic semantic augmentation strategy, SCINet not only learns the features of the data but also captures its semantic meanings. This enables the model to excel in handling rare or ambiguous categories, improving its ability to generalize and manage the complexity and diversity of the real world. 

\bibliographystyle{IEEEtran}
\bibliography{Reference}


\end{document}